%% file: main.tex
\documentclass[runningheads]{llncs}
\usepackage{graphicx}
\usepackage{amsmath,amssymb} 
\usepackage{color}
\usepackage[dvipsnames,svgnames,x11names]{xcolor}
\usepackage{soul}
\usepackage{fancyhdr}

\usepackage[export]{adjustbox}
\usepackage{xr}

\usepackage{caption}
\usepackage{subcaption}
\usepackage{amsmath}
\usepackage{amssymb}

\usepackage[finalold]{trackchanges}

\newcommand{\js}[1]{\textcolor{black}{#1}}
\newcommand{\ice}[1]{\textcolor{black}{#1}}

\graphicspath{{./Images/}}

\begin{document}

\newcommand{\point}{
    \raise0.7ex\hbox{.}
    }


\pagestyle{headings}
\mainmatter

\title{Lost in Time: Temporal Analytics for Long-term Video Surveillance} 
\titlerunning{Temporal Analytics for Long-term Video Surveillance}

\authorrunning{Khor \& See}
\author{Huai-Qian Khor \and John See}
\institute{Center for Visual Computing, Faculty of Computing and Informatics\\Multimedia University, 63100 Cyberjaya, Malaysia\\
\email{hqkhor95@gmail.com,johnsee@mmu.edu.my}
}

\maketitle
\thispagestyle{empty}

\begin{abstract}
Video surveillance is a well researched area of study with substantial work done in the aspects of object detection, tracking and behavior analysis. With the abundance of video data captured over a long period of time, \js{we can understand patterns in human behavior} and scene dynamics through data-driven \emph{temporal analytics}. 
In this work, we propose two schemes to perform descriptive and predictive analytics on long-term video surveillance data. We generate heatmap and footmap visualizations to describe spatially pooled trajectory patterns with respect to time and location. We also present two approaches for anomaly prediction at the day-level granularity: a trajectory-based statistical approach, and a time-series based approach. Experimentation with one year data from a single camera demonstrates the ability to uncover interesting insights about the scene and to predict anomalies reasonably well.

\end{abstract}

\input{Introduction}
\input{relatedWork}
\input{proposedFramework}

\input{Methods}

\input{Experimentation}

\input{Discussion}
\input{Conclusion}

\bibliographystyle{splncs}
\bibliography{references}

\end{document}

%% file: Introduction.tex
\section{Introduction}

In the domain of video surveillance, there has been a significant amount of research done in the past few decades relating to a variety of sub-tasks such as object detection and tracking \cite{zhang2015new}, and behavior analysis \cite{morris2013understanding}. The abundance of video data in the ``Big Data" era has resulted in far more data collected than analysed or processed~\cite{porikli2013video}. 
Across a long period of time, \emph{temporal analytics} can offer interesting data-driven insights into a variety of contemporary problems such as retail location analysis, and understanding of commuting behaviors and crowd patterns. 


The Long-term Observation of Scenes with Tracks (LOST) dataset \cite{IEEE:LOST} is the only known dataset established for the purpose of studying scene behavior and changes at a longer time scale. Its data consists of videos captured from a number of outdoor streaming cameras located in different parts of the world, over a period of 1--3 years. Each video contains the same half an hour period captured each day. The dataset comes with rich metadata (i.e. blob, trajectory) that can be readily used for further analysis of long term trends in the scene.

The analysis of long term trends has been much studied in a wide variety of fields such as climatology \cite{collins2013long} and epidemiology \cite{velagaleti2008long}. Within the engineering and computing domains, long term trend analysis has also been investigated in the area of time series forecasting \cite{timeseries} and social media analytics \cite{twitter}. While video surveillance research has progressed tremendously in many aspects, most of the data considered are short term in nature. Moreover, data analytics applied to long-term surveillance data could potentially derive a deeper understanding into changes that occur at a longer time scale (months to years). A few recent works have begun exploring the usefulness of long-term video data for anomaly mining and prediction \cite{IEEE:monday,IEEE:lostworld,hu2015detection}.



In this paper, we describe the notion of \emph{temporal analytics} and \js{how it is carried out to extract valuable insights from long-term video surveillance data}. \js{The main contribution of this work is to propose feasible techniques for descriptive and predictive analytics} on video surveillance data that spans a long period of time. Firstly, we performed a descriptive extraction of trajectory patterns from a monitored scene to generate \emph{heatmap} and \emph{footmap}, which can capture time- and location-based trends pooled from the accumulated trajectories. Secondly, we proposed two approaches for predicting anomalies at the \emph{day-level granularity}: a trajectory-based statistical approach \ice{ that calculates statistical difference with distance between trajectories } and a time series-based classification approach \ice{ that computes statistical difference in terms of number of trajectories per day.} 
\js{The statistical approach is motivated by the work in \cite{lostandfound} which used trajectory-based information to predict abnormal trajectories, while time-series based approach allows daily trajectory information to be represented temporally.}
We report the insights obtained from these proposed schemes by experimenting with one-year data from a single camera of the LOST dataset.

%% file: relatedWork.tex
\begin{figure*}[t]
\centering
\includegraphics[scale=0.22]{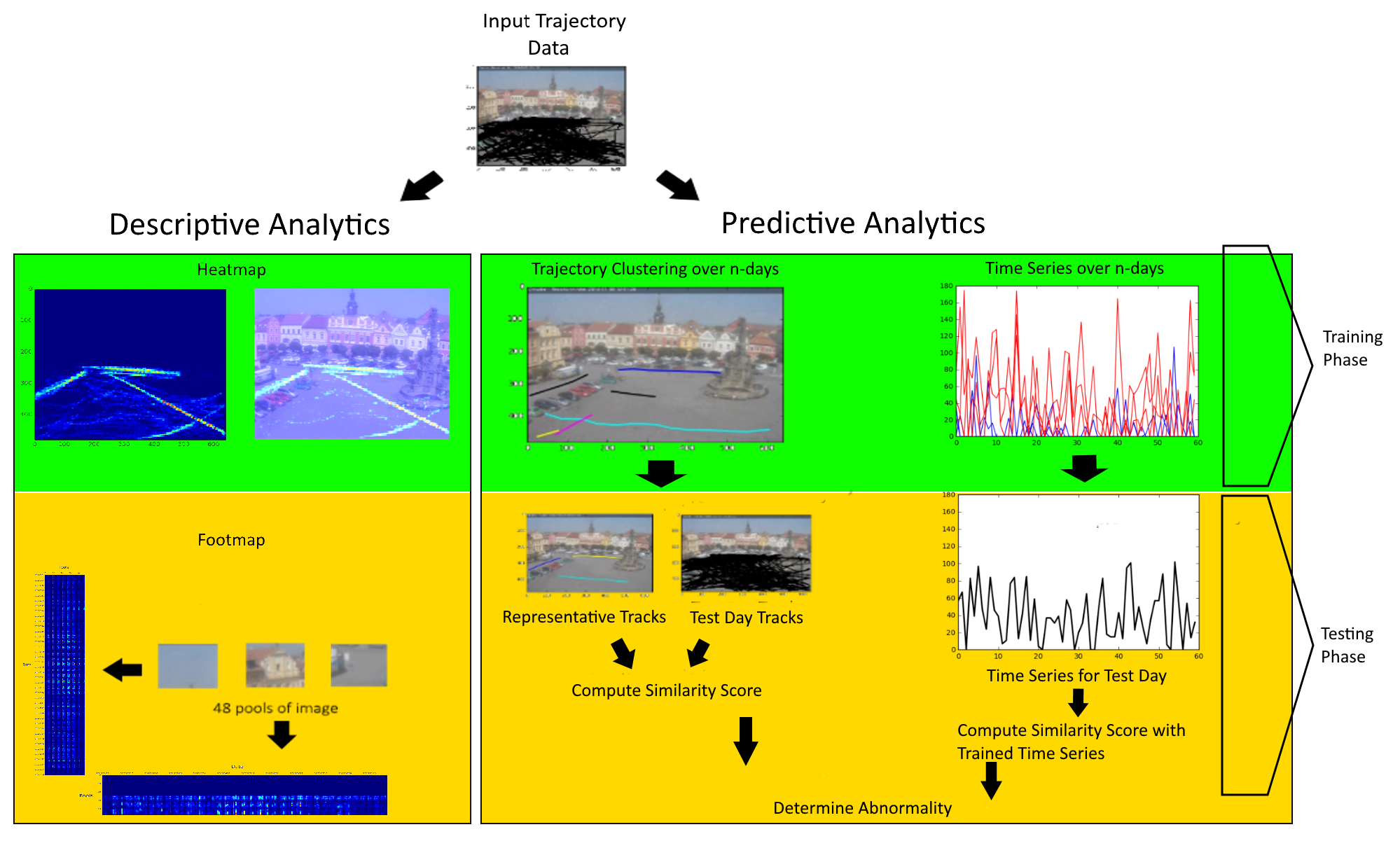}
\caption{Proposed temporal analytics framework for long-term video surveillance}
\label{fig:framework}
\vspace{-1em}
\end{figure*}

\section{Related Work}
Long-term video surveillance research is still in a nascent stage. Recently, there has been an increase of interest in investigating how common behavioral patterns can be mined and how anomalies can be predicted across longer time spans.

In the original work that proposed the LOST dataset, Abrams et al. \cite{IEEE:LOST} demonstrated a few example cases of how trajectories of moving objects (or ``tracks'') can be clustered and aggregated into statistics that effectively captures the long-term trends in track variation over the scene. They also showed how these statistics also correlated to external signals such as day of the week and weather condition. 

Following that, a work by See and Tan \cite{IEEE:lostworld} proposed a time-scale based framework for mining anomalous track patterns on selected cameras from the LOST dataset. Their method first clusters unlabeled tracks using two temporal levels to find common modes of behavior, represented by track exemplars. Then, a probabilistic anomaly prediction algorithm was devised to evaluate the abnormality of new tracks. Due to the lack of ground truth labels, prediction of abnormal tracks was performed using synthetically generated anomalous tracks. A subsequent work \cite{lostandfound} performed object classification on objects extracted from LOST videos for more than 23,000 frames under a variety of weather conditions.

Zen et al.~\cite{IEEE:monday} proposed a pixel-wise approach to determine the density of traffic captured for a whole month in New York City. The segmented foreground regions of moving objects were extracted to compute traffic density. Typical patterns
and anomalous events were discovered by an anomaly
score, which defines the distance between the traffic densities taken at a specific day of the week and time of the day. The vulnerability of this method is that the presence of noise as foreground pixels can produce inaccurate traffic patterns.

%% file: proposedFramework.tex
\section{Data Preparation}
In this work, we use the object trajectory data from camera number `001' (Ressel Square, Chrudim,
Czech Republic) of the LOST dataset \cite{IEEE:LOST}, spanning 228 days between 01-01-2012 to 31-12-2012. In this period of 1 year, a portion of videos had a lower frame rate than most of the other videos while some had erroneous trajectory data. These videos were omitted as they were found to be unsuitable for our analysis. The trajectory data contains essential information of the moving objects in the scene, i.e. track ID and object centroid coordinates and dimensions.

One pressing issue is the lack of annotations in the original dataset \cite{IEEE:LOST}, which is essential for validating the predictive analytics task. Hence, we sought the help of three annotators to manually annotate the selected videos with a binary anomaly label (i.e. '1' corresponds to an anomalous day while '0' corresponds to a normal day). The annotators are asked to provide labels independently (no knowledge of labels given by the others), and the final annotated label is decided on the basis of consensus where two or more annotators agree to the same annotation. The criteria for a day to be considered as an anomaly are: 1) The occurrence of ad-hoc events at the plaza, 2) The occurrence of vehicles driving through the plaza which is only meant for pedestrians only.

%% file: Methods.tex
\section{Framework \& Methods}
In this section, we present our proposed temporal analytics framework, and the methods associated to the two schemes -- descriptive and predictive analytics. Fig. \ref{fig:framework} shows a graphical illustration of the proposed framework, outlining how trajectory data can be visualized and used for the purpose of day-granularity anomaly prediction. The purpose of selecting day-level granularity is to experiment with a coarser granularity that spans a longer period of time.


\begin{figure*}[!t]
\centering
\begin{subfigure}[t]{0.36\linewidth}
	\includegraphics[width=\textwidth]{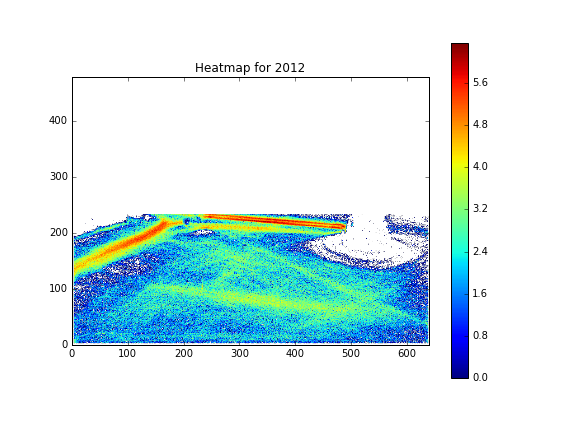}
    \vspace{-1.5\baselineskip}
    \caption{ }
	\label{fig:heatmap}
\end{subfigure}
\hspace{-1.5em}
\begin{subfigure}[t]{0.34\linewidth}
	\includegraphics[width=\textwidth]{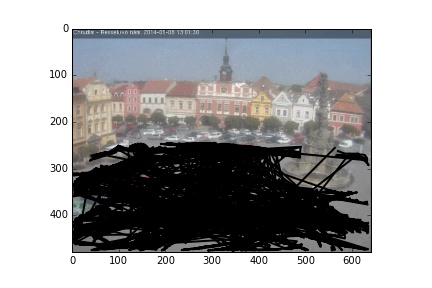}
    \vspace{-1.5\baselineskip}
    \caption{ }
	\label{fig:alltracks}
\end{subfigure}
\hspace{-1.5em}
\begin{subfigure}[t]{0.34\linewidth}
	\includegraphics[width=\textwidth]{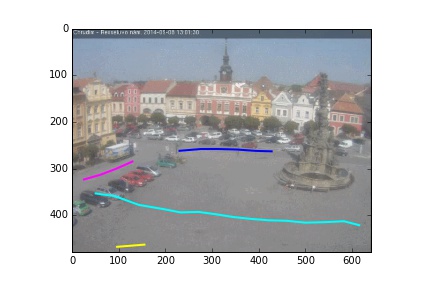}
    \vspace{-1.5\baselineskip}
	\caption{ }
	\label{fig:cluster}
\end{subfigure}
\vspace{-0.25em}
\caption{(a) Heatmap constructed from a period of one year (January-December 2012) for camera 001, (b) All tracks accumulated over a span of three similar days, (c) Representative tracks of clusters, each shown with a distinct color.}
\vspace{-1em}
\end{figure*}

\subsection{Trajectory Data}

The initial data contains trajectories (or tracks, in short), which are each represented by a set of points in 2D space. Each track $T$ is denoted as 
\begin{equation}
	t_i = \{x_{i}, y_{i}\} \qquad \forall i \in 1...|T|
\end{equation}
where $x$ and $y$ are the track coordinates at frame $i$.

\subsection{Descriptive Analytics}
The aim of performing \emph{descriptive analytics}~\cite{bigdata} 
on long-term surveillance data is to derive useful knowledge from historical data that can potentially be used for further analysis. In our work, descriptive analytics can be performed to extract higher level information such as the busiest period in the year or busiest day in a usual week, or paths that most objects will pass through. The outcome of this module is two-fold -- we utilize \emph{heatmaps} to discover the paths that are most commonly taken by moving objects. Besides, we also introduce a new visualization called \textit{footmap} that is able to summarize the intensity of activity in the monitored scene based on time and location.

Heatmaps \cite{fisher2007hotmap} are constructed by accumulating the track coordinates $t(x,y)$ "traveled" by the moving objects, 
\begin{equation}
	\mathcal{H} = \sum_{k}^{|K|} \sum_{i}^{|T_{k}|} \delta_{t_{i}} \quad \forall t=t_{i}
	\label{eq:heatmap}
\end{equation}
for $K$ number of tracks per day, over a period of $N$ days.

Since the traveled locations are largely centered on a small number of prominent paths, we used a logarithmic scaled heatmap to provide a better balance in the color intensity distribution in the scene. Figure \ref{fig:heatmap} shows the generated heatmap of moving objects represented by an array of different colors. Blueish regions represent lesser activities while the reddish regions represent higher level of activities. In this 1-year heatmap of camera `001' taken from the LOST dataset, we observe that the two main thoroughfares are distinctly marked in red, while two pedestrian pathways at the plaza are in noticeable green streaks. 

To extract information on the intensity of activity based on time and location, we create a new type of visualization called a \textit{footmap} for the monitored scene. Footmap can be used to capture time- and location-based trends pooled from patches of accumulated trajectories.
\textit{Footmap} is constructed by first accumulating all trajectory points $t$ within each $ 80 \times 80 $ non-overlapping square patch, also called a \textit{pool}, by sum operation. 
To ensure all patches in a scene are of same size, we select the patch size based on two criteria: 1) it must be a common divisor of both the width and height of the scene; 2) it is of a reasonable size to allow the difference in the intensity of activity to be clearly depicted without being too homogeneous (patches too small) or too coarse (patches too large).

\begin{figure}[t!]
\centering
    \vspace{-11\baselineskip}
	\includegraphics[width=1.05\linewidth]{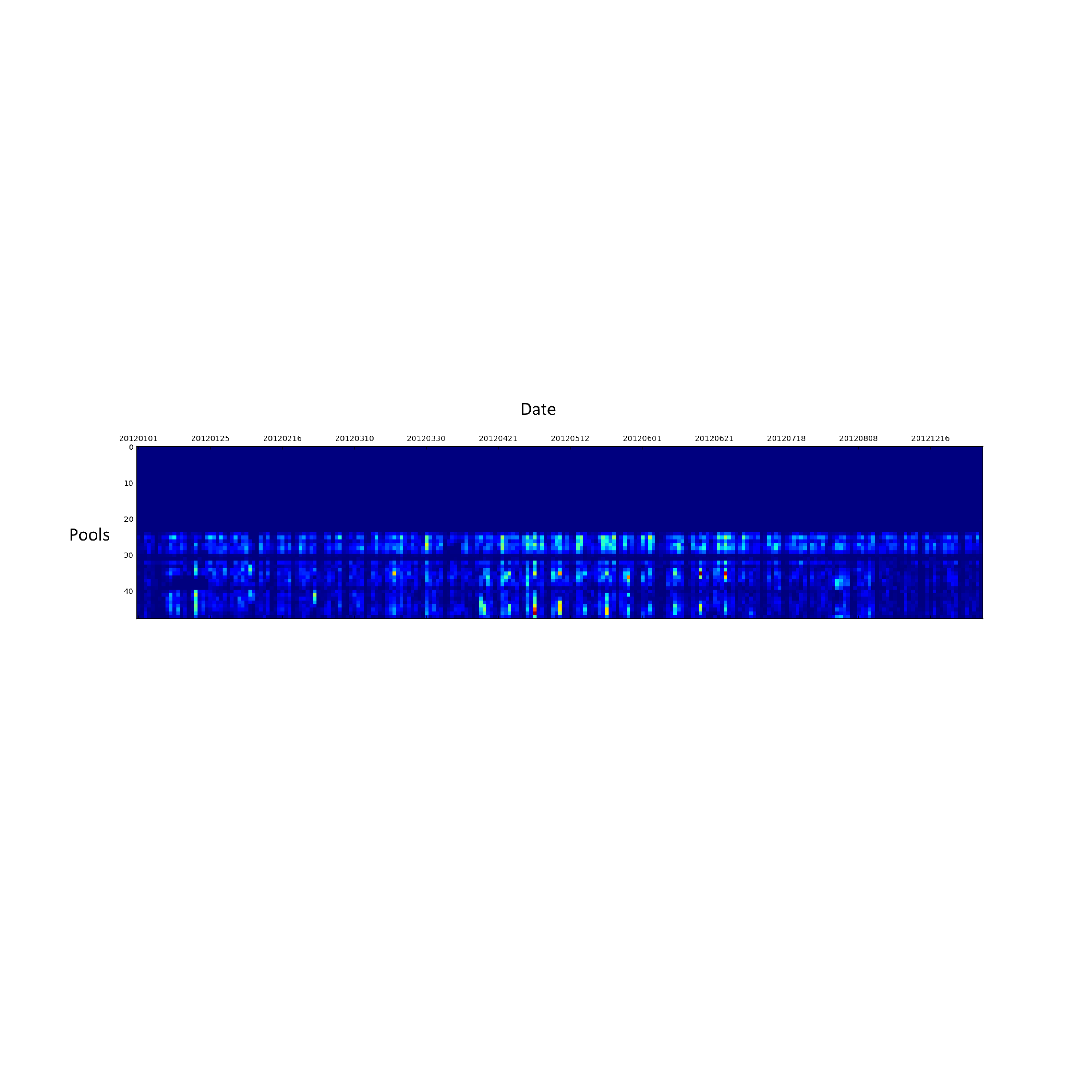}

    \vspace{-14\baselineskip}
	\caption{Horizontal Footmap (HF)}
    \label{fig:footmap1}
\end{figure}

\ice{We name the footmap as a Horizontal 
Footmap (HF) (in Figure \ref{fig:footmap1}). }
The HF is generated by spatially pooling all pixels in each patch before rearranging the output values vertically, taking from the original scene in left-to-right row-wise fashion. This causes the patches from the top parts of the scene to be located near the top of the footmap. This is repeated for all days in the period of analysis; we use a period of 1 year from January to December 2012. 

The intuition behind the use of footmap is the visualization of activities from the perspective of both duration and location. Using a typical jet colormap (red for high intensity, blue for low intensity), we observe that the footmap demonstrates the intensity of activities across different times of the year (i.e. date axis), and also across different locations in scene (i.e. pool numbers). The spring/summer season, which is near the middle zone, is noticeably busier particularly around April to June. In terms of location, the road area is visibly busy on most days of the year (middle rows of the HF) while there are obviously no activities detected in the top portion of the scene that corresponds to the buildings and sky (top half rows of the HF).


\subsection{Predictive Analytics}

We propose two strategies for \emph{predictive analytics} on long-term surveillance data: the first being trajectory based while the second is time-series based. 

\subsubsection{Trajectory based Prediction}

In this method, we first detect for anomalies at a finer \emph{trajectory-level} granularity. This is then used to make an inference at the \emph{day-level} granularity, whether a particular day is likely to be abnormal, relative to what usually happens on that same day of the week (Mondays, Tuesdays, ..., Sundays). Hence, this assumes that each day exhibits a consistent pattern of activities throughout the year.

First step involves clustering the trajectories $T$ to obtain common modes of movements in the scene. We utilize a trajectory-specific clustering algorithm, TraClus \cite{IEEE:traclus} to perform clustering on all tracks in the scene. The clustering is done on a time-scale of $(\omega, \epsilon)=(28, 7)$ as defined in \cite{IEEE:lostworld}; a time-scale defines a temporal window that spans a specific number of days ($\omega$) at a specific stride
($\epsilon$) or periodicity. This can be interpreted as taking 3 previous similar days (e.g. Mondays) as training days for cluster generation while the 4th similar day is used as the test day for anomaly prediction. Figure \ref{fig:alltracks} shows all tracks from the three training days, which are then grouped by TraClus algorithm into four clusters, each represented by its representative track (centroid of cluster) in Figure \ref{fig:cluster}.

Motivated by previous works \cite{IEEE:hu,IEEE:lostworld} that apply a statistical approach for probabilistic anomaly prediction, we utilize a similar concept to that, extending it further to day granularity anomalies. We formulate the distance metric, 
\begin{equation}
D(P,Q) = \frac{1}{|P|} \sum_{t_{p} \in P} \min_{t_{q} \in Q} |t_{p} - t_{q}|^2
\label{equation:distance}
\end{equation}
as a random variable that measures the distance between new track and clusters.

Given the computed distance between a test track $T'$ and the $j$-th cluster containing the representative track, we approximate the likelihood probability,  
\begin{equation}
P(T'|X_j) = e ^ {-n_{j}D(T', X_j)}
\label{equation:probability}
\end{equation}
The parameter $\eta_j$ can be computed based on maximal likelihood evaluation, which is the reciprocal of mean distances learnt from the distribution of $K$ training tracks:
\begin{equation}
\eta_j = \frac{K}{\sum_{k=1}^{K} D(T_k, X_j)}
\label{equation:etaj}
\end{equation}
For ease of computation, we compute the thresholds $\Gamma = \{\gamma_1, \gamma_2, ..., \gamma_J\}$ for all $J$ clusters during the training process. Each threshold $\gamma_j$ is calculated by taking the minimum likelihood of all training tracks $T_{j,k}$ belonging to the $j$-th cluster,
\begin{equation}
\gamma_j = \min_{k} P(T_{j,k} | X_j)
\label{equation:gamma}
\end{equation}
where a test track $T'$ is classified as an anomalous track if the likelihood of the test track given cluster $X_j$,
\begin{align}
j^* = \arg\max_{j} P(T'|X_j)
\label{equation:j_threshold}
\end{align}
is less than its respective threshold $\gamma_j^*$ and the distance between the test track and the nearest cluster $D(T'|X_{j^{*}}) > \delta$, an empirically defined distance threshold; we use $\delta=1000$.

To predict a day-level anomaly on a test day $A$, we compute the ratio of number of anomalous tracks over the total number of tracks on that day, $\psi_T' = N_{ano}/N_{total}$. If $\psi_T$ is more than a empirically defined anomaly threshold $\lambda$, then the test day will be predicted as anomalous.
\begin{equation}
A = 
	\begin{cases} 
	1 \quad \text{if }\psi_T < \lambda \cr			
	0 \quad \text{otherwise}
	\end{cases}
\label{equation:day-anomaly}
\end{equation}

\subsubsection{Time series based Prediction}

The second strategy performs anomaly prediction using time series trajectory data. Data can be summarized for each day, by counting the unique number of active trajectories (i.e. moving objects) within a defined interval $\theta$ seconds. For instance, a 30-minute video with interval $\theta=15$ seconds would produce a time series of 120 values. We denote the time-series count data as, 
\begin{equation}
	C(s) = \sum_{s=0}^{S}[O(s)]
\label{eq:timeseries}
\end{equation}
where $S=D*60/\theta$ and $O(s)$ represent the unique trajectories occuring at time interval $[\theta s,\theta (s+1)]$.

To measure the similarity between two time series, we opt for dynamic time warping (DTW) which seeks to find the optimal non-linear alignment. To speed things up, we use the Keogh lower bound (LB) variant of DTW (famously known as `LB\_Keogh') \cite{lbkeogh} which computes in linear time. With this quick method to determine the similarity between two time series, we predict an anomaly at the day-level using the classic k-Nearest Neighbor algorithm with LB\_Keogh as distance measure. For every time series in the test set, a search is performed through all points in the training set to match with the most similar time series in the training set. We applied a 50:50 training-test split on a total of 212 days that contain valid time series data; the first half data is used to predict the second half data.
  

%% file: Experimentation.tex
\section{Experimental Results and Discussion}
In this section, we discuss the outcome of the two schemes performed on long-term video trajectory data.

\subsection{Descriptive Analysis}
In a year's worth of data, we uncover certain patterns through our descriptive analytics of heatmap and footmap.

\begin{figure*}[!t]
\centering
\begin{subfigure}[t]{0.3\linewidth}
	\includegraphics[width=\textwidth]{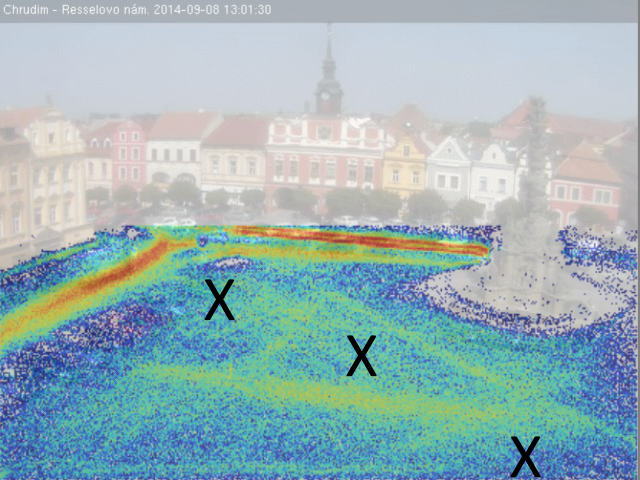}
    \vspace{-1.25\baselineskip}
    \caption{ }
	\label{fig:xmarks}
\end{subfigure}
\begin{subfigure}[t]{0.435\linewidth}
	\includegraphics[width=\textwidth]{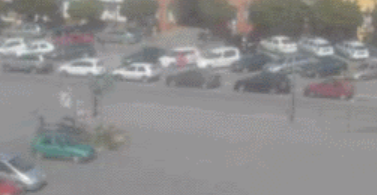}
    \vspace{-1.25\baselineskip}
	\caption{ }
	\label{fig:middlepool}
\end{subfigure}
\vspace{-0.25em}
\caption{(a) The X-marks denote suitable places for marketplace stalls or advertisements 
(b) Two patches that have the highest amount of activity throughout the year, i.e. the 27th and 28th pool of the HF (c.f. Figure \ref{fig:footmap1}).}
\vspace{-1em}
\end{figure*}

The first analysis that we present is the outcome of heatmap. By overlaying the heatmap on top of the scene as in Figure \ref{fig:xmarks}, we can see that a reasonable amount of moving objects, possibly pedestrians and crowds, occurred at the plaza area. Hence, future developments such as marketplace stalls or advertisement billboards can be strategically placed at potential locations, adjacent to the paths that are mostly used by pedestrians (shown with X marks in Figure \ref{fig:xmarks}).

The second analysis is done based on the footmap in Figures \ref{fig:footmap1}. From the Horizontal Footmap, we observe the difference in color intensities around April--June. This is indicative of an increase in activities due to more movements, hence retail activities can be suggested to increase during this busy period.

\subsection{Anomaly Prediction}

One of the main issues that we faced in measuring the performance of predicting anomalies is the imbalanced number of typical (negative) and anomalous (positive) days. As such, the standard accuracy metric is less suitable, and may not reflect the actual performance. Hence, we obtain the full confusion matrix, which allows us to determine the Precision, Recall and F1-score measures. 

For trajectory-based prediction, we report the best F1-score of 0.46 based on the threshold $\lambda=0.01$. Meanwhile, the time series-based method appears to perform much better at predicting anomalies, achieving the best F1-score of 0.67 with DTW window size of 2 ($k=1$ therefore 1-NN). This demonstrates the robustness of using the time-series data for predicting anomalies.

\begin{table}[!t]
\centering
\caption{Confusion Matrix for Trajectory based Prediction}
\begin{tabular}{|l|c|c|} 
\hline
 \textbf{Predicted} $\backslash$ \textbf{Desired}  & Anomalous & Typical \\ 
 \hline
 Anomalous & 50 & 110 \\ 
 \hline
 Typical & 7 & 46 \\
 \hline
\end{tabular}
\label{table:confusion}
\vspace{-1em}
\end{table}

\begin{table}[!t]
\centering
\caption{Confusion Matrix for Time Series based Prediction}
\begin{tabular}{|l|c|c|} 
\hline
 \textbf{Predicted} $\backslash$ \textbf{Desired}  & Anomalous & Typical \\ 
 \hline
 Anomalous & 13 & 23 \\ 
 \hline
 Typical & 10 & 60 \\
 \hline
\end{tabular}
\label{table:confusion2}
\vspace{-1em}
\end{table}

\begin{table}[!t]
\centering
\caption{Performance Results of Various Metrics for Anomaly Prediction}
\begin{tabular}{ |c|c|c| } 
\hline
\textbf{Metric} & \textbf{Trajectory} & \textbf{Time-Series} \\
\hline
Precision & 0.31 & 0.67 \\
\hline
Recall & 0.88 & 0.69\\
\hline
F1-Score & \textbf{0.46} & \textbf{0.67}\\
\hline
\end{tabular}
\label{table:scores}
\vspace{-1em}
\end{table}

\ice{Table \ref{table:confusion} and \ref{table:confusion2} are confusion matrices which display the number of predicted anomalies for trajectory-based and time-series-based methods respectively. The metrics used in Table \ref{table:scores} are described as follows: Precision is the number of correct results divided by the total number of returned results, Recall is the number of correct results divided by the number of results that should have been returned, F1-score is the harmonic mean of precision and recall which often used for class-imbalanced data. }

\subsection{Discussion}
\ice{To the best of our knowledge, there are no prior works related to anomaly prediction on a day-level granularity for this dataset due to the lack of ground truth labels. Hence, it is not possible to compare directly in terms of standard performance measures. Prior works \cite{IEEE:lostworld,IEEE:crowdanomaly} that focus on trajectory-level prediction showed some promising results but no ground truth labels were available for validation. Interestingly, using this coarser granularity, we show that the visualization of traffic patterns can yield a distinction between high and normal traffic flow from both the temporal (Figure \ref{fig:footmap1}) and spatial (Figure \ref{fig:heatmap}) perspectives.}

%% file: Conclusion.tex
\section{Conclusion}
In this work, we present a framework for performing temporal analytics for long-term video surveillance; it consists of a descriptive extraction of trajectory patterns to generate useful visualizations, and two predictive schemes for identifying anomalies at the day-level granularity. 
\ice{This is a preliminary attempt at proposing descriptive and predictive analytics on long-spanning temporal information from surveillance videos. There is still plenty of room for improvements on the techniques proposed, and how object trajectories can be better represented with additional directional information.}
\js{We hypothesize that temporal analytics on long-term video data will have far-reaching benefits 
for various domains such as urban planning, market strategy for businesses, and public security.}

